\documentclass[conference]{IEEEtran}

\IEEEoverridecommandlockouts
\usepackage{cite}
\usepackage{amsmath,amssymb,amsfonts}
\usepackage{algorithmic}
\usepackage{graphicx}
\usepackage{textcomp}
\usepackage{xcolor}
\usepackage[utf8x]{inputenc}
\usepackage{fonts-tlwg}

\def\BibTeX{{\rm B\kern-.05em{\sc i\kern-.025em b}\kern-.08em
    T\kern-.1667em\lower.7ex\hbox{E}\kern-.125emX}}

\begin{document}

\title{Coffee Roast Intelligence}

\author{\IEEEauthorblockN{Sakdipat Ontoum\IEEEauthorrefmark{1}\IEEEauthorrefmark{2},
Thitaree Khemanantakul\IEEEauthorrefmark{1}\IEEEauthorrefmark{2}, \\ Pornphat Sroison\IEEEauthorrefmark{1}\IEEEauthorrefmark{2}
Tuul Triyason\IEEEauthorrefmark{1}\IEEEauthorrefmark{2}, Bunthit Watanapa\IEEEauthorrefmark{1}\IEEEauthorrefmark{2}}
\IEEEauthorblockA{\IEEEauthorrefmark{1}Computer Science Program, School of Information Technology}
\IEEEauthorblockA{\IEEEauthorrefmark{2}King Mongkut’s University of Technology Thonburi}
\IEEEauthorblockA{Bangmod, Thung-Khru, Bangkok, Thailand}}

\maketitle

\begin{abstract}
As the coffee industry has grown, there would be more demand for roasted coffee beans, as well as increased rivalry for selling coffee and attracting customers. As the flavor   of each variety of coffee is dependent on the   degree of roasting of the coffee beans, it is vital to maintain a consistent quality related to the degree of roasting. Each barista has their own method for determining the degree of roasting. However, extrinsic circumstances such as light, fatigue, and other factors may alter their judgment. As a result, the quality of the coffee cannot be controlled. The “Coffee Roast Intelligence” application is a machine learning-based study of roasted coffee bean degrees classification  produced as an Android application platform that identifies the color of coffee beans by photographing or uploading them while roasting. This application displays the text showing at what level the coffee beans have been roasted, as well as informs the percent chance of class prediction to the consumers. Users may also keep track of the result of the predictions related to the roasting level of coffee beans.

\end{abstract}

\begin{IEEEkeywords}
Roasted Coffee Beans, Coffee Beans degree of Roasting, Machine Learning, Classification, Android Application
\end{IEEEkeywords}

\section{Introduction}
Coffee is now the world’s most popular beverage, with over 400 billion cups being consumed each year [1]. Furthermore, the coffee industry has expanded around the world, especially in Thailand. Coffee consumption in Thailand has been continuously growing year after year. As a result, Thailand’s coffee shop industry has expanded in recent years. According to statistics, the global coffee market is expected to reach 60,000 million baht in 2020, up by 10.7\% from the previous year. Thai coffee bean exports totals to  81,000 tons, making it the world’s tenth largest coffee exporter [2]. In comparison to 2016-2017, there is a 4\% rise [5]. As a result, the popularity of coffee consumption among the new generation is growing and will continue to expand. Thais are predicted to consume more than 300,000 tons of coffee each year in the next five years, and this number is expected to rise [3].

The demand for roasted coffee beans will increase as the coffee industry grows. Roasted coffee beans come in a variety of degrees, depending on the temperature at which they were roasted [3]. The degree to which coffee beans are roasted is the most essential aspect that defines the taste of the coffee in the cup. The taste of coffee varies depending on the degree to which it has been roasted. The user inspects physical quality features such as color and form to classify the coffee beans according to their degree of roasting. However, environmental factors such as light, fatigue, and other factors might alter human judgment [4].

As a result, the notion of creating a system that uses technology and roasted coffee beans to help define degrees of roasting was thought of. The Convolutional Neural Network is the technology employed in this system. A Convolutional Neural Network (CNN) is a component of a computer system that mimics how the human brain analyzes and processes data [5]. Artificial intelligence (AI) is built on this basis, and it solves issues that humans would find impossible or difficult to solve. The Convolutional Neural Network (CNN) is well suited to pattern classification, especially when the classification limitations are not well-defined [6].

\section{Objective}
\begin{itemize}
\item To determine the degree of roasting of coffee beans. 
\item To ensure that the quality of roasted coffee beans is consistent and that the degree of roast is correct.
\item To create a model for classifying the degree of roasting coffee beans based on the image.

\end{itemize}

\section{Scope}
The project, dubbed  “Coffee Roast Intelligence,” is a machine learning-based study of roasted coffee bean classification that has been built as an Android application platform to give baristas and coffee shop owners more ease and precision while roasting beans.

The main goal of this application is to classify the color and form of coffee beans. During roasting, the color of the coffee beans changes from green (unroasted) to darker browns (dark roast). Users can capture or upload an image of the coffee beans while they are being roasted. This application will classify the degree of roasting by presenting the degree of roasting in textual form, as well as the probability percentage of class prediction, using image processing and artificial neural networks. Users will be able to make better decisions if the accuracy of determining the coffee roasting degree is steady. The application also has a function that allows users to save the prediction of their coffee bean roasting level. A description can be added to the coffee beans that the user has photographed or uploaded. This feature aids consumers in recalling details about their favorite coffee beans.

\section{literature review}
\subsection{Classification of Coffee Bean Degree of Roast Using Image Processing and Neural Network [7]}
The research uses image processing and artificial neural networks to classify the degree of roast to reduce errors and ensure accuracy in determining the roast level of a coffee bean. Only one coffee variety, Excelsa, and one coffee origin were used to develop the process (Indang, Cavite). A smartphone was used to snap a photo of roasted Excelsa coffee beans. Furthermore, an artificial neural network was used to identify the degree of roasting of coffee beans into light roast, medium roast, and extremely dark roast utilizing RGB values as input.

\subsection{An Intelligent System for Coffee Grading and Disease Identification [8]}
This thesis has an intelligent method for assessing coffee and classifying its diseases. It shows models created with deep learning approaches that help farmers and coffee grading professionals make better decisions. The paradigm for grading raw coffee beans into twelve quality levels is being developed. Despite the fact that the dataset is tiny in comparison to the quantity of data necessary to train deep learning models, a decent model (89.1\%) accuracy on the test dataset could be achieved, outperforming both conventional machine learning techniques and off-the-shelf deep learning architectures.

\section{dataset}
Roasted coffee beans has been roasted at JJ Mall Jatujak’s, “Bona Coffee.” There are four roasting levels. The green or unroasted coffee beans are Laos Typica Bolaven (Coffea arabica). Laos Typica Bolaven is the light  roasted coffee bean (Coffea Arabica). Doi Chaang (Coffea Arabica) is medium roasted, whereas Brazil Cerrado is dark roasted (Coffea Arabica) [9].

The coffee bean photos are captured with an IPhone12 Mini with a 12 megapixel back camera, Ultra-wide and Wide Camera. The camera is set at a location with a plane parallel to the object’s path when photographs are being captured. Images of roasted coffee beans are captured in a variety of settings  to validate a wide range of roasted coffee bean image inputs. This experiment employs both LED light from a light box and natural light to shoot the dataset; then the image’s noise is enhanced by putting each variety of coffee bean in a container. Images are automatically collected and saved in PNG format. Each example bean’s image is 3024x4032 pixels in size. Each example bean’s image is 3024x4032 pixels in size. There are 4800 photos in total, classified in 4 degrees of roasting. There are 1200 photos under each degree. The sample dataset is shown in Figures 1–4.

\begin{figure}[htbp]
\centerline{\includegraphics[scale=0.1]{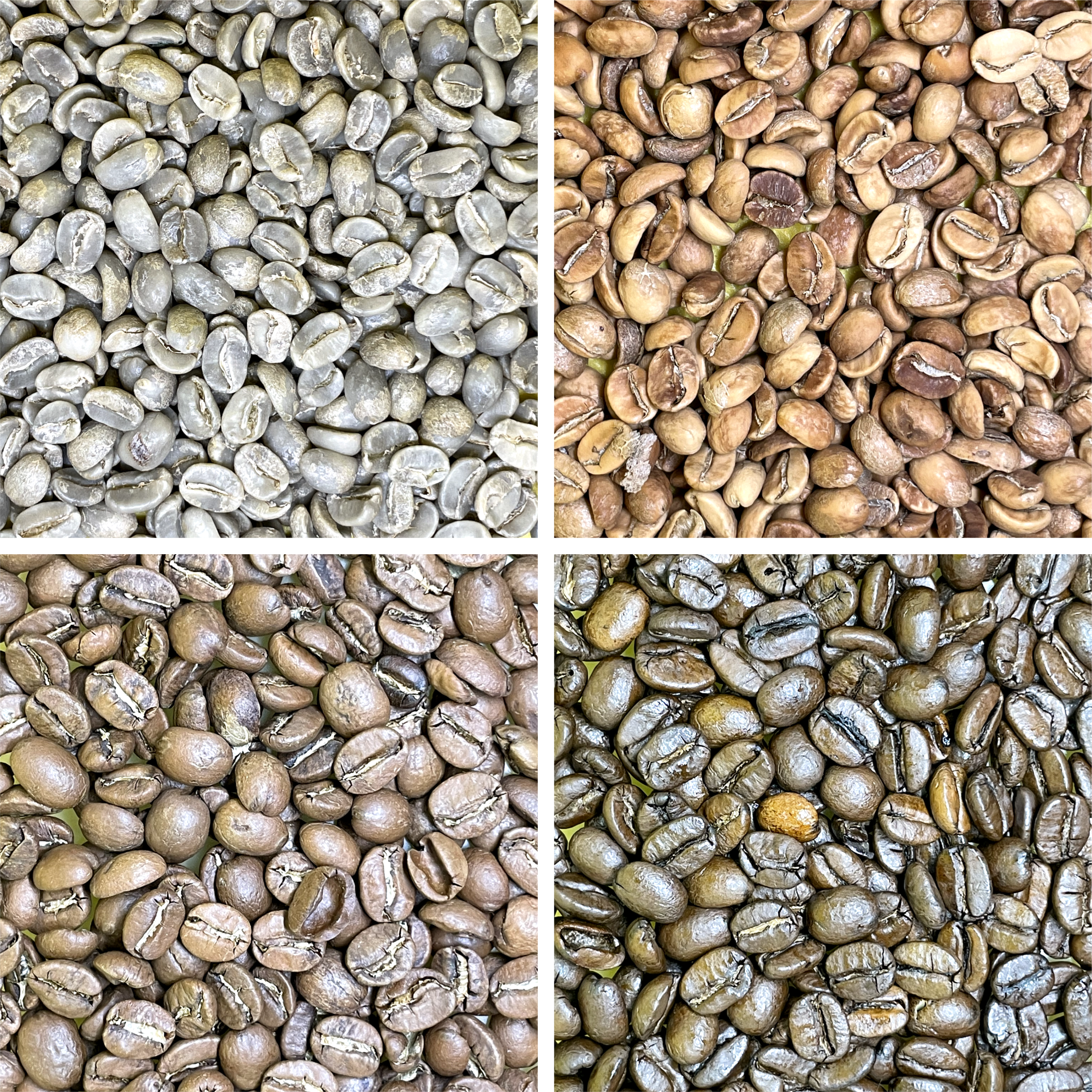}}
\caption{Represents an example dataset of group of coffee beans photographed in lightbox.}
\end{figure}

\begin{figure}[htbp]
\centerline{\includegraphics[scale=0.1]{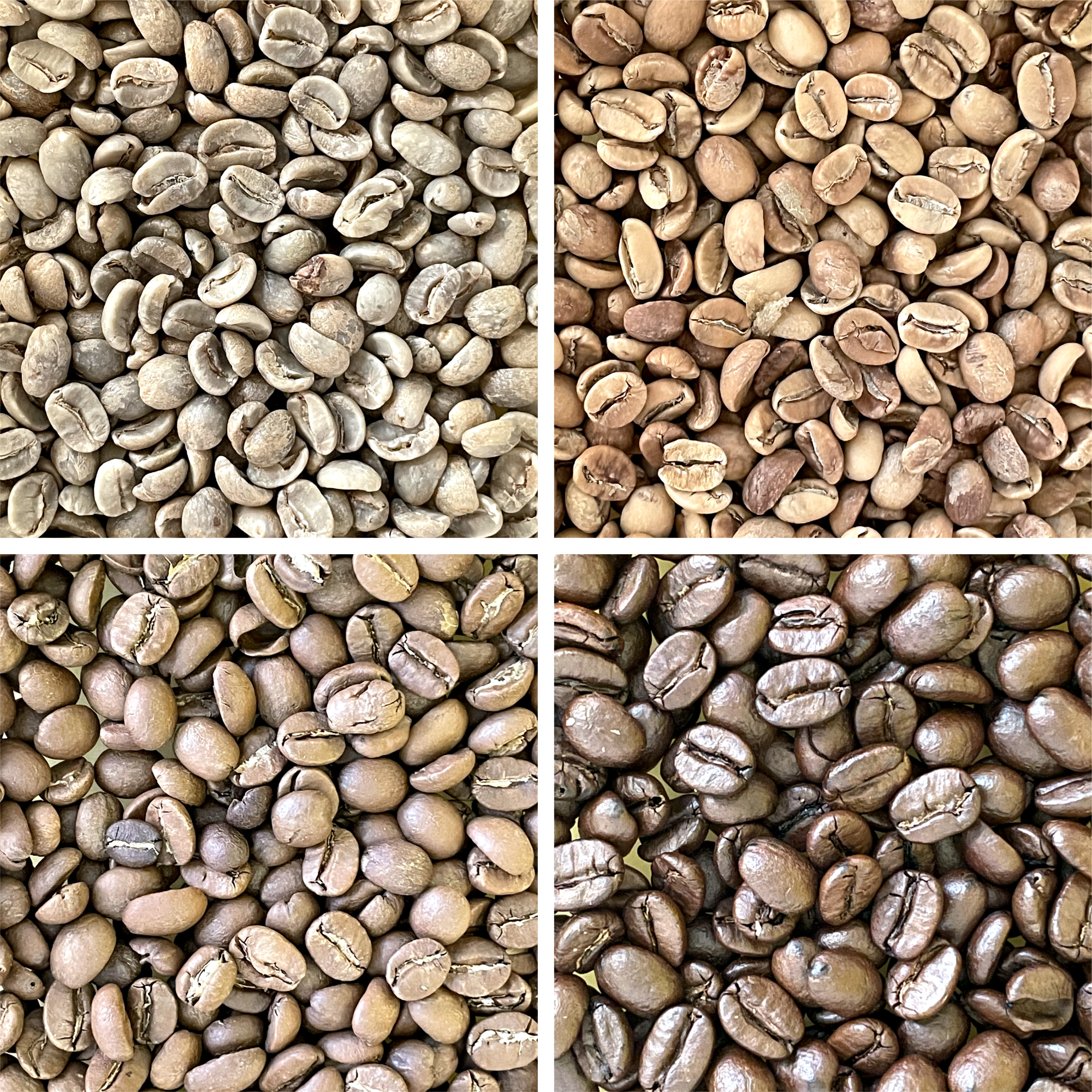}}
\caption{Represents an example dataset of group of coffee beans photographed
in natural light.}
\end{figure}

\begin{figure}[htbp]
\centerline{\includegraphics[scale=0.1]{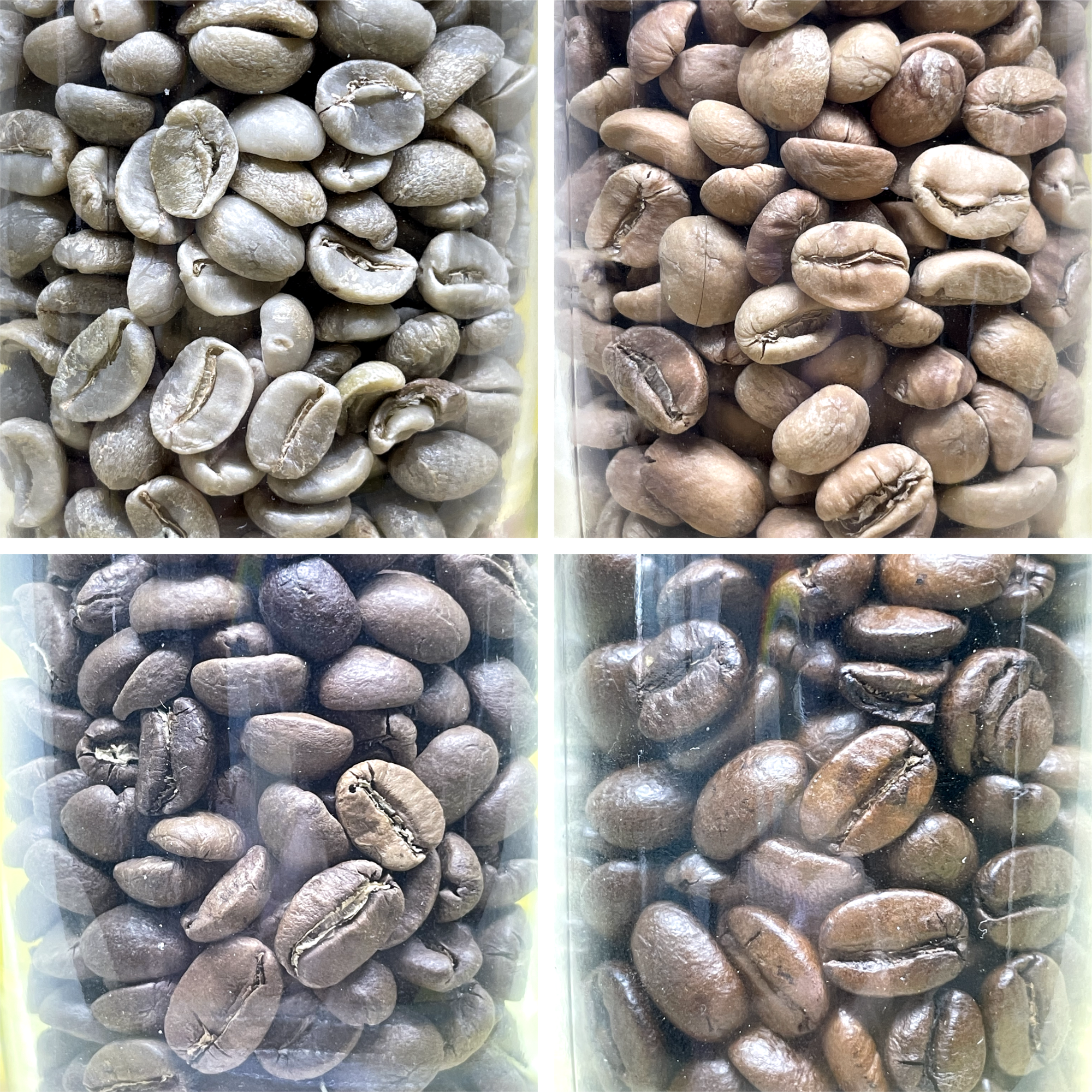}}
\caption{Represents an example dataset of group of coffee beans photographed in lightbox and glass bottle.}
\end{figure}

\begin{figure}[htbp]
\centerline{\includegraphics[scale=0.1]{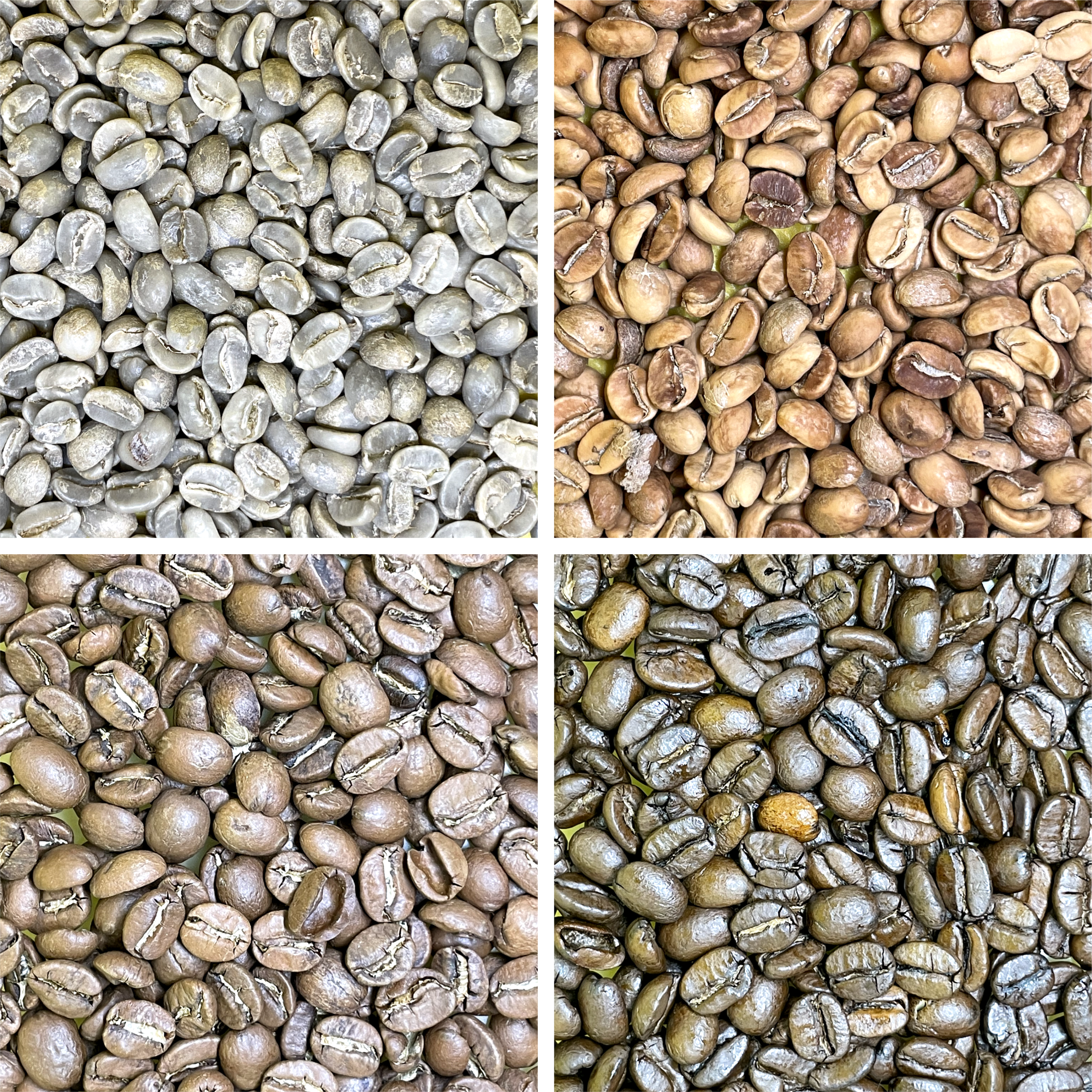}}
\caption{Represents an example dataset of group of coffee beans photographed in natural light and glass bottle. }
\end{figure}

\section{methodology}
After shooting the roasted coffee beans, this project gathered the entire dataset on Google Drive to make it easy to grab datasets for model training. To assess how accurate the coffee roast classification model is, a machine learning model is constructed using Google Colab Notebook, which is based on the Python programming language and employs Keras and TensorFlow to train and test datasets. Then machine learning is applied to the Hierarchical Data Format version 5 (hdf5.file), an open-source file format for massive, complicated, and heterogeneous data [10].

\subsection{Image Preprocessing}
The purpose of image pre-processing is to improve the image and make it more helpful for modeling. To begin with, the data image is cleaned from the original by using gaussian blur to remove noise. After that, the RGB image is transformed to HSV (Hue, Saturation, and Value), which is a color scheme based on Hue, Saturation, and Value [11]. A mask and a Boolean mask are also created to decide whether an image pixel should be treated. The image is not processed if a pixel with a value of 0 is turned black. If the pixel value is greater than or equal to 0 (value 1), the image can be processed. Fig.5. shows an example of image preparation.

\begin{figure}[htbp]
\centerline{\includegraphics[scale=0.35]{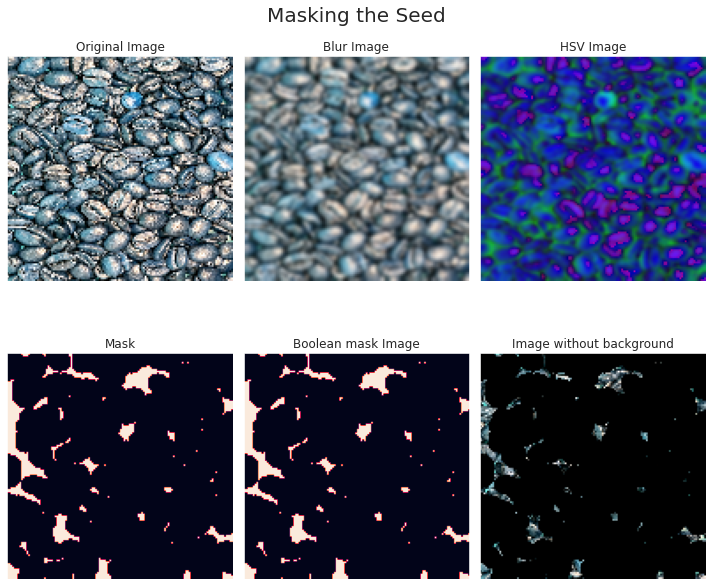}}
\caption{Represents an example image preprocessing that use gaussian blur, convert color to HSV, create Boolean mask and apply to get image without background.}
\end{figure}

After that, the image is normalized by changing the input values from [0...255] to [0...1], because the Convolutional Neural Network trains better with [0...1] input. Finally, the image is enhanced by rotating, zooming, shifting, and flipping it so that the Machine Learning model can see many perspectives of the image during training.

\begin{figure}[htbp]
\centerline{\includegraphics[scale=0.35]{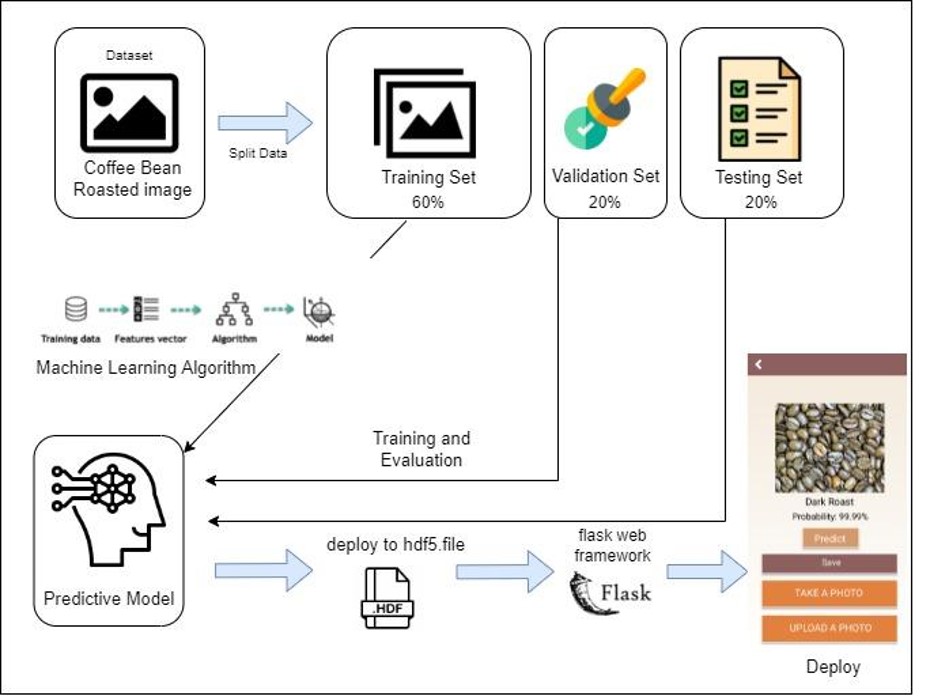}}
\caption{Machine Learning Workflow Diagram of the system}
\end{figure}

\subsection{Model Implementation}
The Python programming language, Keras, and TensorFlow,
an open-source toolkit for training and testing models, are used in the development of this project. The datasets for coffee beans have been preprocessed, normalized, and augmented. The whole Dataset will be divided into three sections: training 60\%, validation 20\%, and test 20\%. Following that, machine learning is trained and applied to the Hierarchical Data Format version 5 (hdf5.file), show in Fig. 6.

\subsection{Machine Learning Model}
This project employs transfer learning to create a machine learning model for the Mobilenet model [12], which is a pretrained model with an input layer of 244 x 244 pixels, which corresponds to the image’s length and breadth respectively. Then, as our own regularization, we add batch normalization and dropout to make the model good at training photographs quickly and reliably, as well as addressing certain overfit issues. The output of a Mobilenet model is as shown in Fig.7.

\begin{figure}[htbp]
\centerline{\includegraphics[scale=0.35]{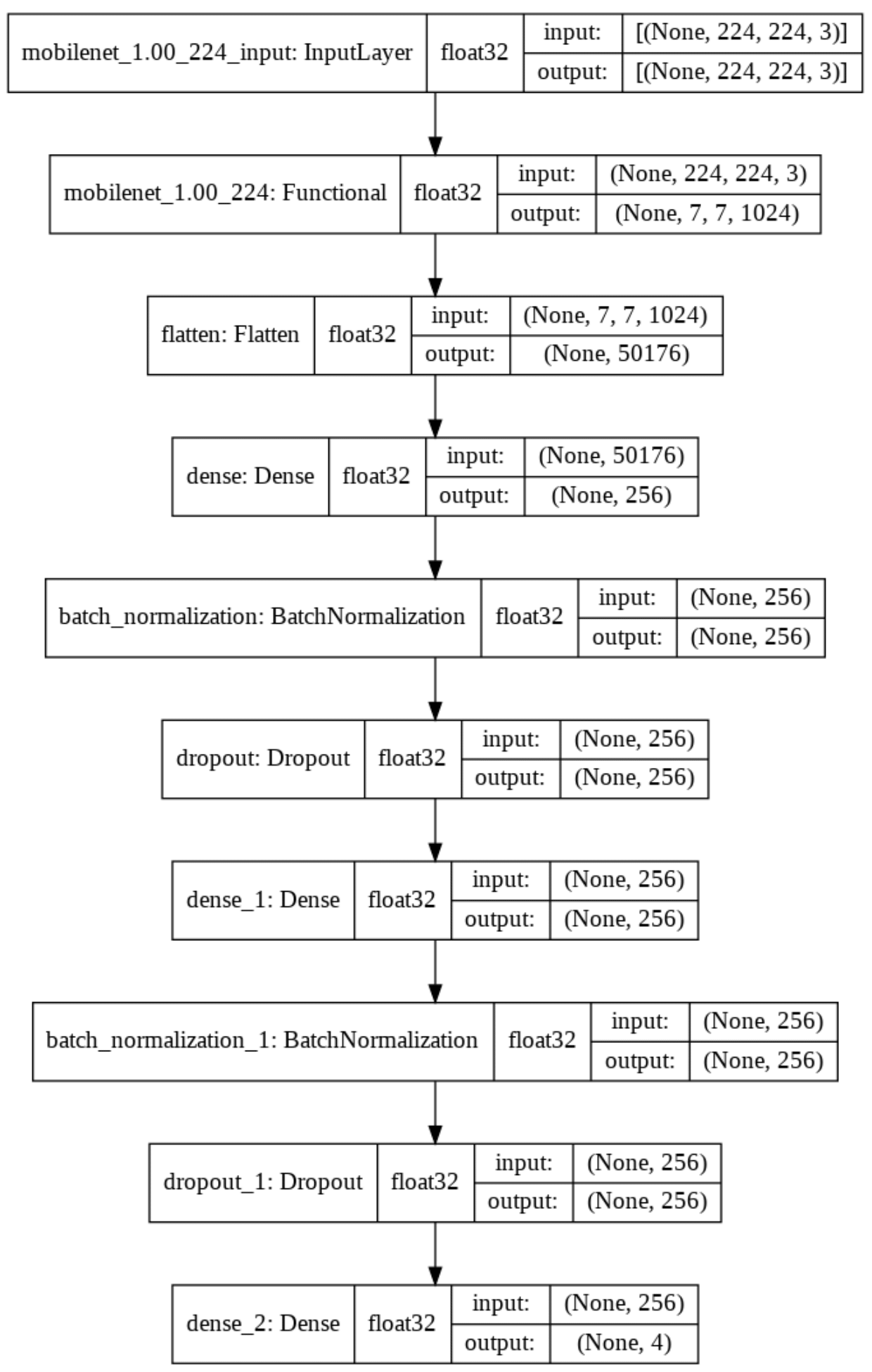}}
\caption{Model Architecture}
\end{figure}

Furthermore, machine learning models employ a technique known as k-fold cross-validation. Cross-validation is a resampling approach used to test machine learning models on a small set of data. When used to make predictions based on data not included during the model’s training, it can increase overall performance [13]. The k-fold process begins with a random shuffle of the dataset. The dataset is then separated into k groups. The number K denotes the number of groups into which a given data sample will be divided. Each group acts as a test, training, or holdout dataset. In the last step to assess the skill of a machine learning model on unseen data, the model’s skill is summed using a sample of model evaluation scores. Table
1. contains more characteristics and information.

\begin{table}[]
\centering
\caption{Parameters and information of training model}
\resizebox{0.55\textwidth}{!}{%
\begin{tabular}{|c|c|c|ll}
\cline{1-3}
\textbf{No.} & \textbf{Parameters} & \textbf{Information}                                                                             &  &  \\ \cline{1-3}
\textbf{1}   & Split Dataset       & \begin{tabular}[c]{@{}c@{}}Trainning 60 \%,\\ Validation 20 \%,\\ and Testing 20 \%\end{tabular} &  &  \\ \cline{1-3}
\textbf{2}   & Target Size         & 224 x 224 pixel                                                                                  &  &  \\ \cline{1-3}
\textbf{3}   & Batch Size          & 32                                                                                               &  &  \\ \cline{1-3}
\textbf{4}   & Function Activation & Relu and Softmax                                                                                 &  &  \\ \cline{1-3}
\textbf{5}   & Optimizer           & Adam                                                                                             &  &  \\ \cline{1-3}
\textbf{6}   & K-fold              & 5                                                                                                &  &  \\ \cline{1-3}
\textbf{7}   & Learning Rate       & 0.00001                                                                                          &  &  \\ \cline{1-3}
\textbf{8}   & Epoch               & 10-40                                                                                            &  &  \\ \cline{1-3}
\end{tabular}%
}
\end{table}

\begin{figure}[htbp]
\centerline{\includegraphics[scale=0.2]{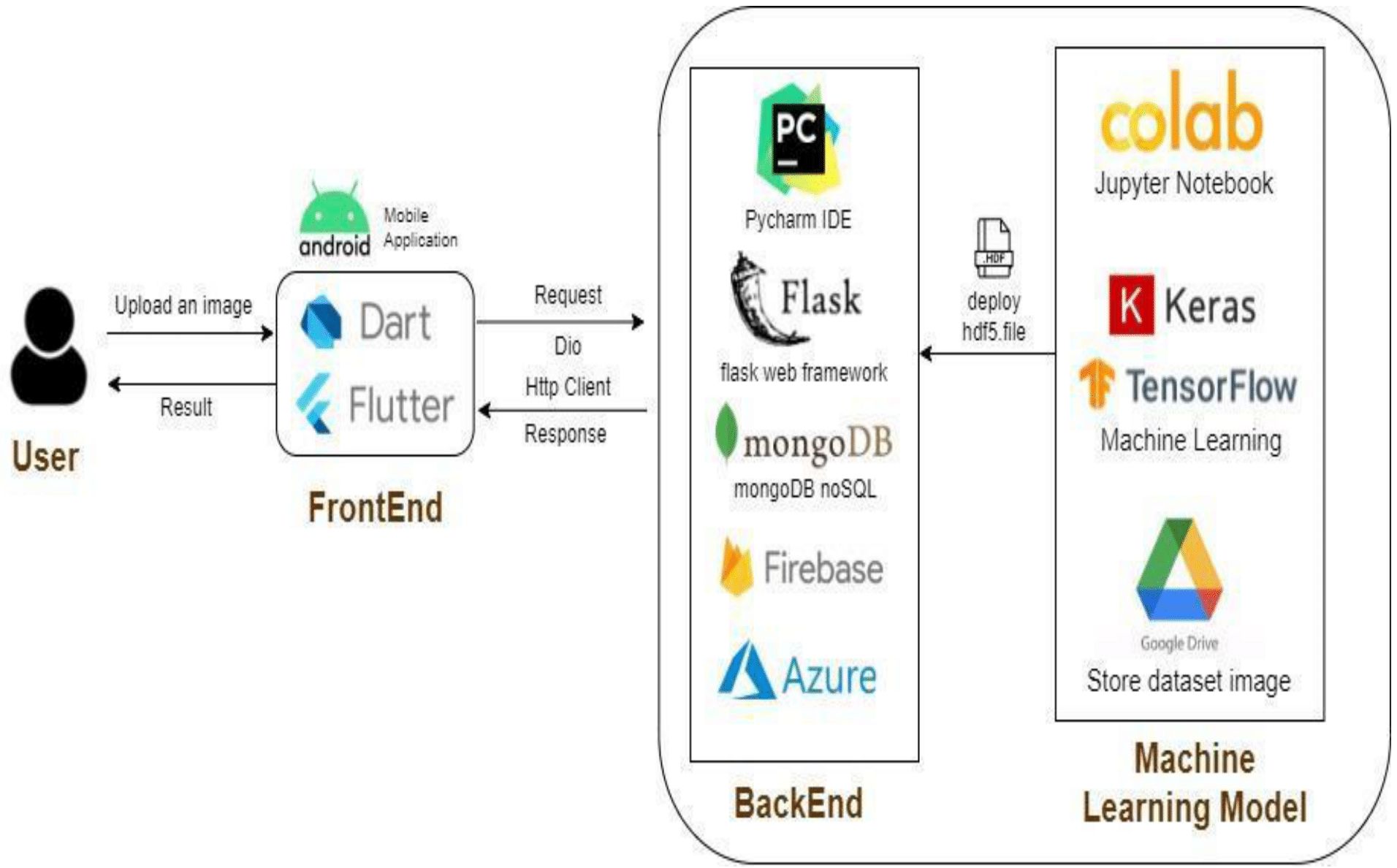}}
\caption{Represents the architecture including the Mobile application, backend, and roasting coffee bean classification machine learning model.}
\end{figure}

\subsection{Back-End}
To deploy the machine learning model to the web server and collecting data into database, we utilize the PyCharm IDE to use the Python flask web framework. It is a code library for creating web applications that are stable, scalable, and maintainable. Firebase is used as a database to collect user data such as name, email, password, and description of the coffee beans image that the user inputs. It is a NoSQL database hosted in the cloud that allows users to store and sync data in real time. MongoDB is used to collect information on the roasting level, percent probability, and timestamp for the display result of coffee beans roasted image. It’s a document- oriented NoSQL database for storing large amounts of data.  In addition, Azure can be used to develop and deploy web applications and APIs on a fully managed platform.

\subsection{Front-End}
To construct an Android application, this project uses the Android Studio tools and the Flutter framework. It’s an open-source UI framework that generates native mobile apps using the Dart programming language. We can refer to the system architecture diagram in Fig.8. for a better understanding of the process.

 \section{Result}
The training-validation accuracy and loss, prediction performance, and confusion matrix of the training, testing, and validation dataset are among the machine learning model findings in this project as shown on Table 2 - 3. The link between accuracies and epochs of each training fold is shown as a graph in Fig.9, where the training-validation accuracy and loss with 5-fold is presented as a graph.

\begin{figure}[htbp]
\centerline{\includegraphics[scale=0.25]{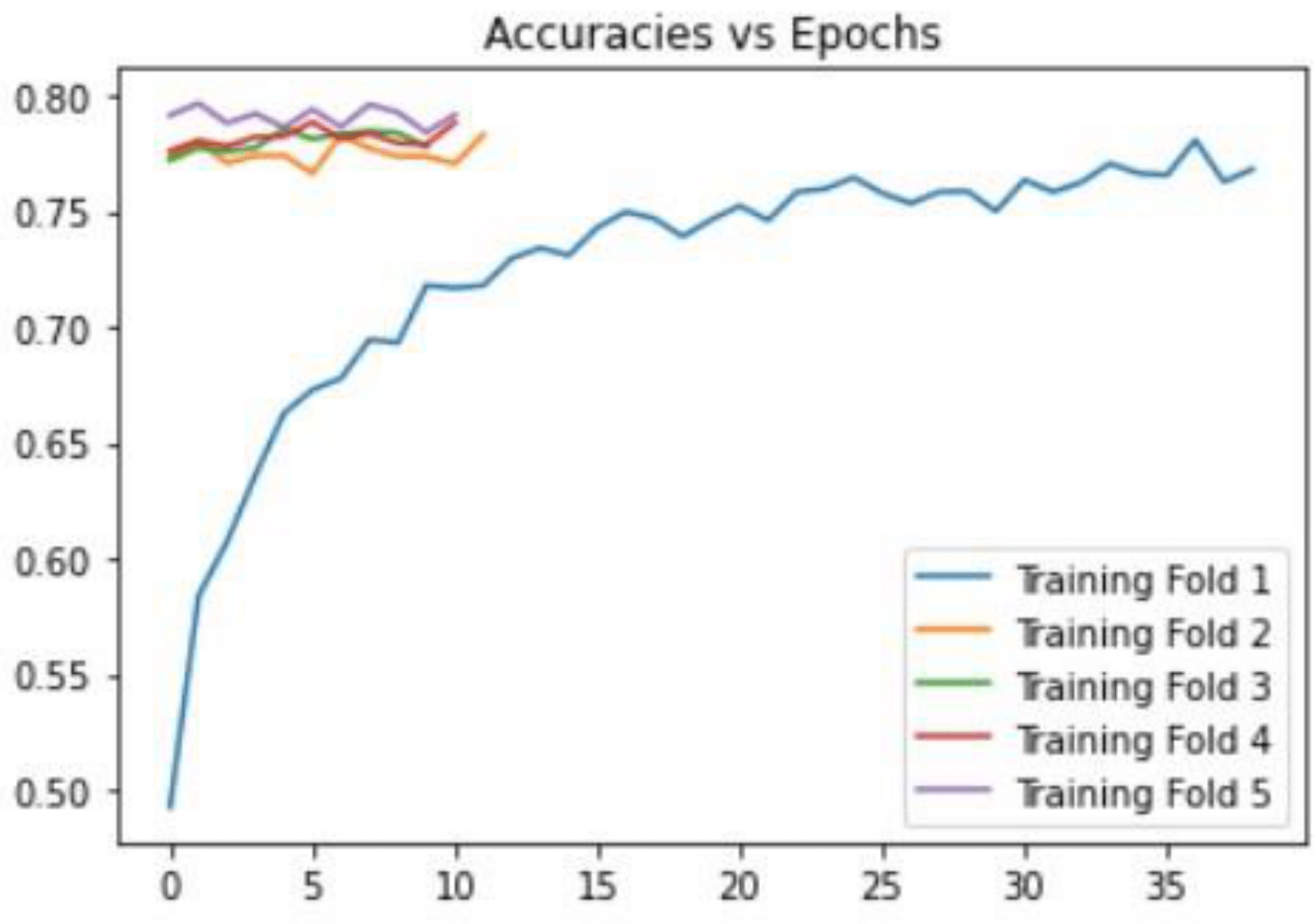}}
\caption{Training-Validation accuracy and loss with each of the training folds.}
\end{figure}

In Fig.10, the accuracy, recall, f1-score, and support scores for each class are presented. The accuracy score is around 0.82, and the majority of the results are between 0.7 and 0.9. 

\begin{figure}[htbp]
\centerline{\includegraphics[scale=0.2]{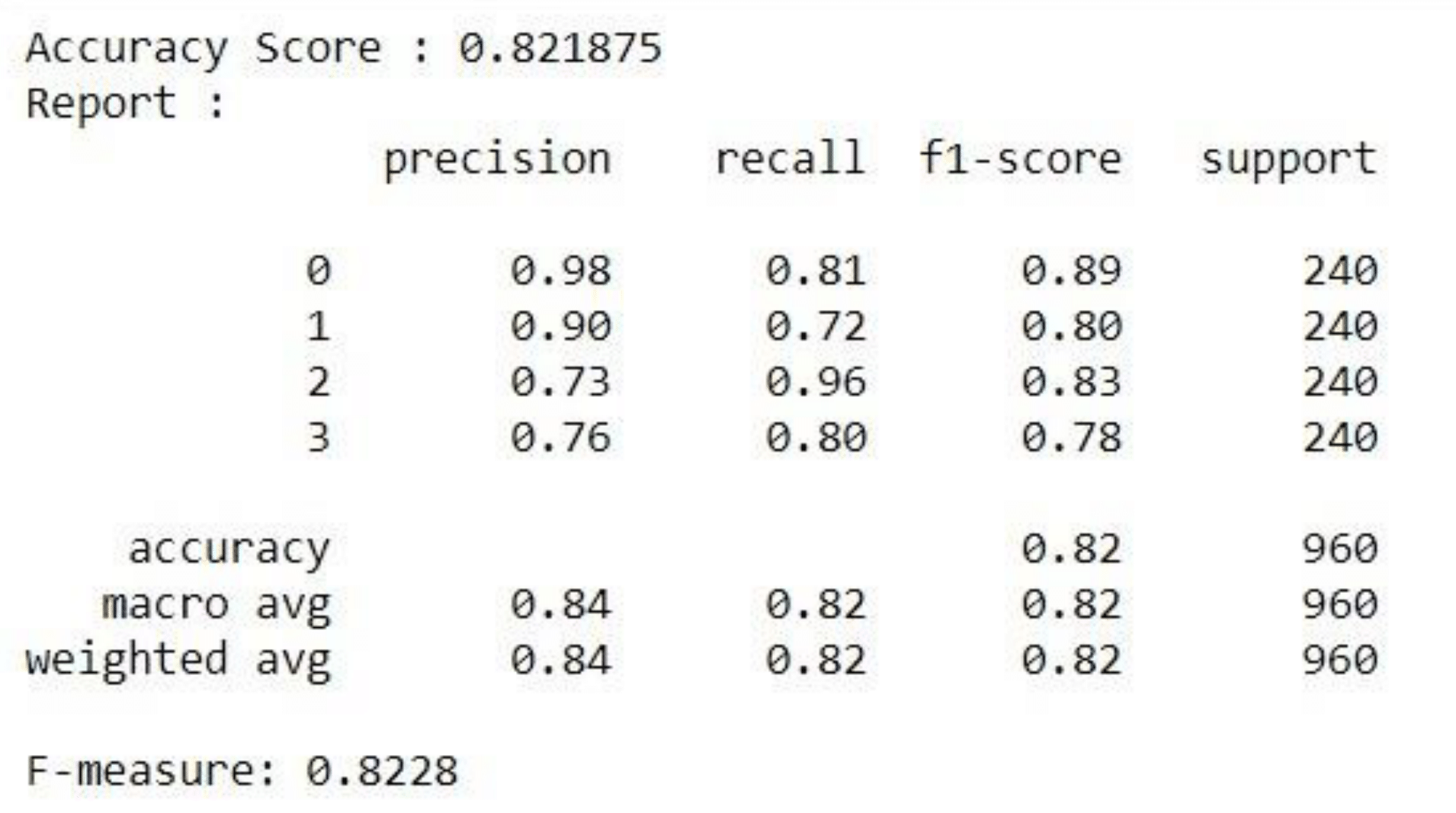}}
\caption{Represents the prediction performance of the Training and Validation dataset of each class. The label class includes Class 0: Dark Roasted, Class   1: Green (Unroasted), Class 2: Light Roasted, Class 3: Medium Roasted.}
\end{figure}

\begin{table}[]
\centering
\caption{Confusion Matrix of Training and Validation dataset}
\resizebox{0.5\textwidth}{!}{%
\begin{tabular}{|c|c|c|c|c|}
\hline
\textbf{Actual/Predict} & \textbf{Green}                      & \textbf{Light}                      & \textbf{Medium}                     & \textbf{Dark}                       \\ \hline
\textbf{Green}          & {\color[HTML]{FE0000} \textbf{194}} & 6                                   & 7                                   & 33                                  \\ \hline
\textbf{Light}          & 2                                   & {\color[HTML]{FE0000} \textbf{172}} & 45                                  & 21                                  \\ \hline
\textbf{Medium}         & 0                                   & 3                                   & {\color[HTML]{FE0000} \textbf{230}} & 7                                   \\ \hline
\textbf{Dark}           & 2                                   & 10                                  & 35                                  & {\color[HTML]{FE0000} \textbf{193}} \\ \hline
\end{tabular}%
}
\end{table}

\begin{table}[]
\centering
\caption{Confusion Matrix of Testing and Validation 
dataset}
\resizebox{0.5\textwidth}{!}{%
\begin{tabular}{|c|c|c|c|c|}
\hline
\textbf{Actual/Predict} & \textbf{Green}                      & \textbf{Light}                      & \textbf{Medium}                     & \textbf{Dark}                       \\ \hline
\textbf{Green}          & {\color[HTML]{FE0000} \textbf{200}} & 5                                   & 54                                  & 21                                  \\ \hline
\textbf{Light}          & 9                                   & {\color[HTML]{FE0000} \textbf{283}} & 8                                   & 0                                   \\ \hline
\textbf{Medium}         & 8                                   & 41                                  & {\color[HTML]{FE0000} \textbf{247}} & 4                                   \\ \hline
\textbf{Dark}           & 4                                   & 5                                   & 45                                  & {\color[HTML]{FE0000} \textbf{243}} \\ \hline
\end{tabular}%
}
\end{table}

This application’s user interface is depicted in Figures 11-13, which includes the homepage, prediction result page, and history page.

\begin{figure}[htbp]
\centerline{\includegraphics[scale=0.25]{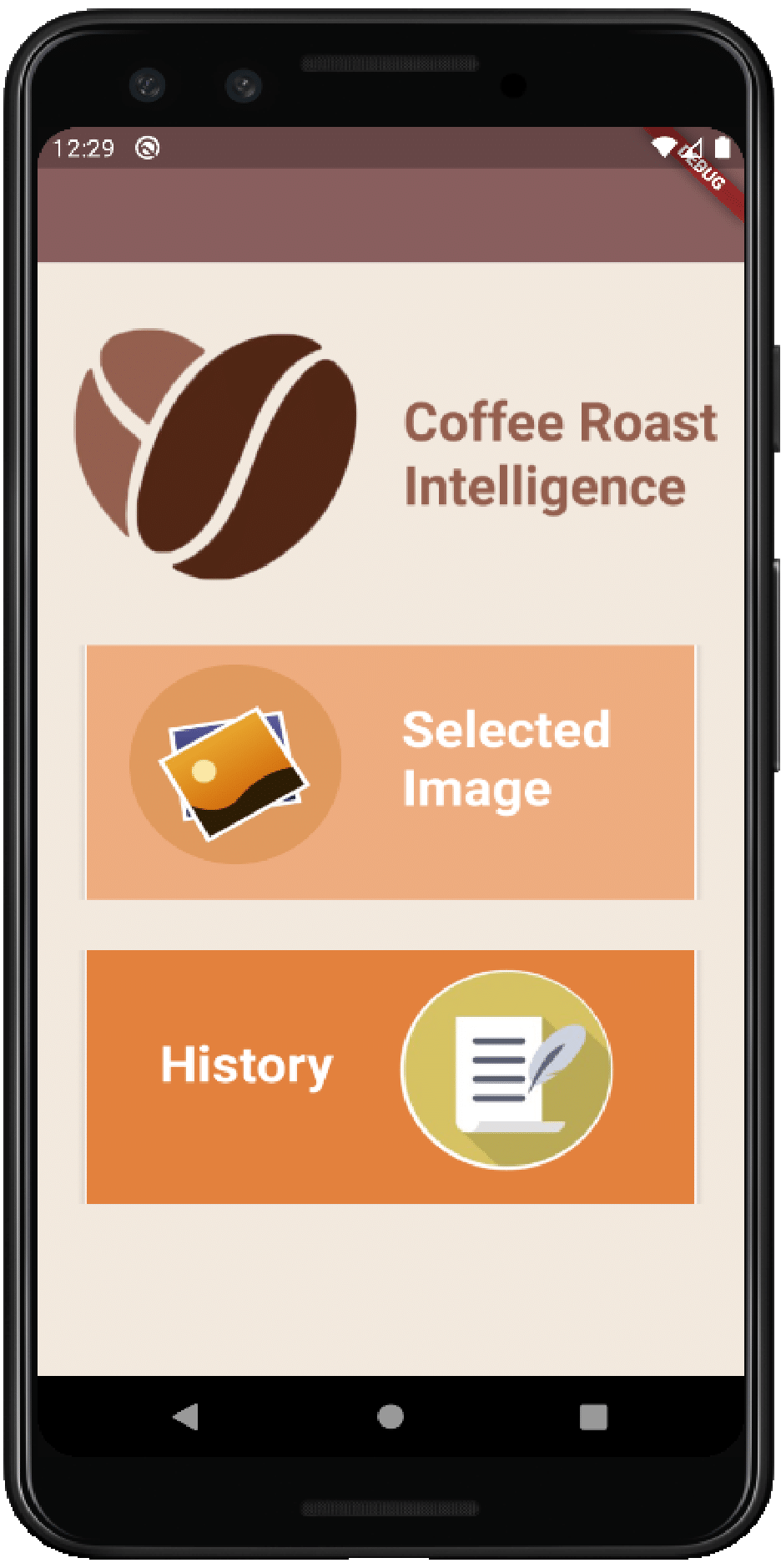}}
\caption{User Home page of the “Coffee Roast Intelligence” mobile application that shows a list of functions of selected image and history.}
\end{figure}

\begin{figure}[htbp]
\centerline{\includegraphics[scale=0.25]{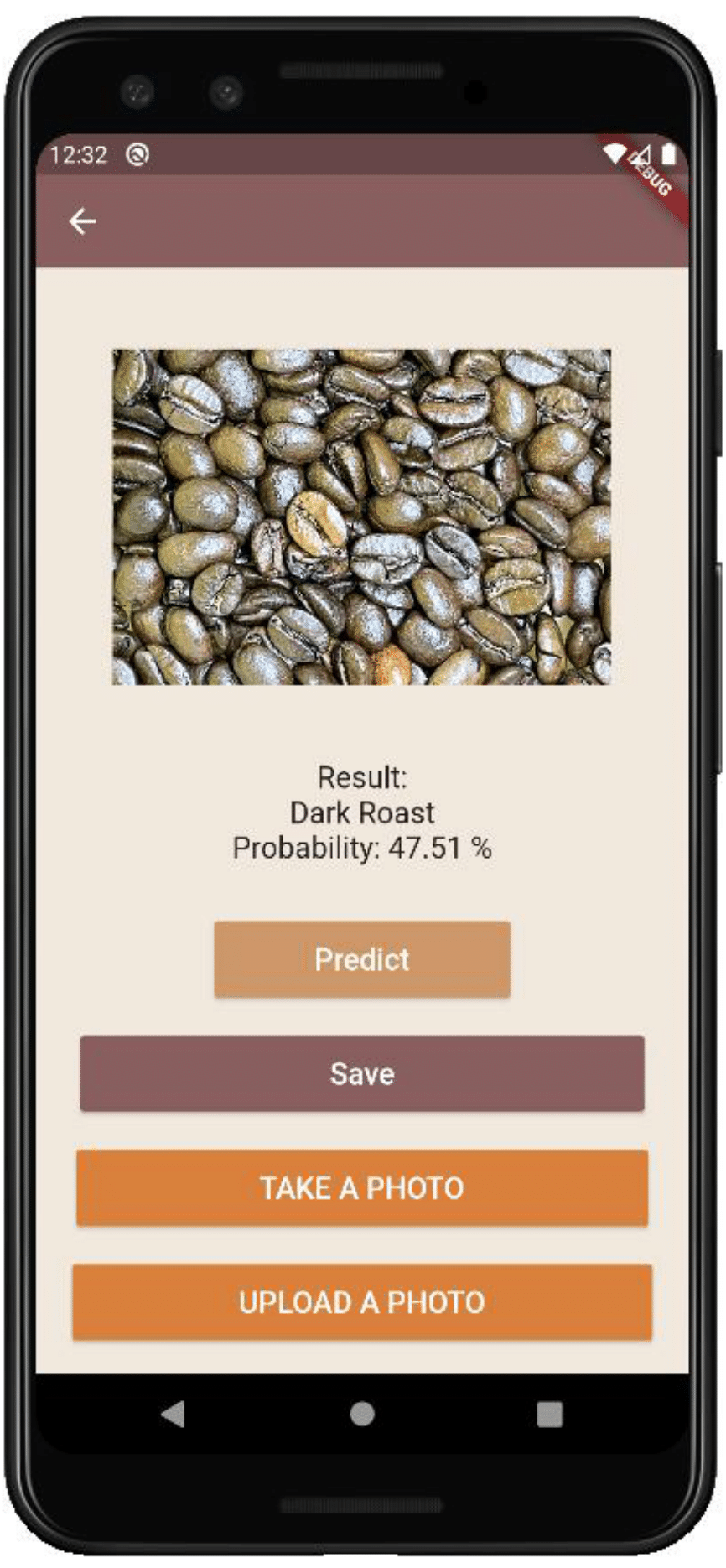}}
\caption{Show results page of the “Coffee Roast Intelligence” mobile application that shows the coffee roasting degree with accuracy.}
\end{figure}

\begin{figure}[htbp]
\centerline{\includegraphics[scale=0.25]{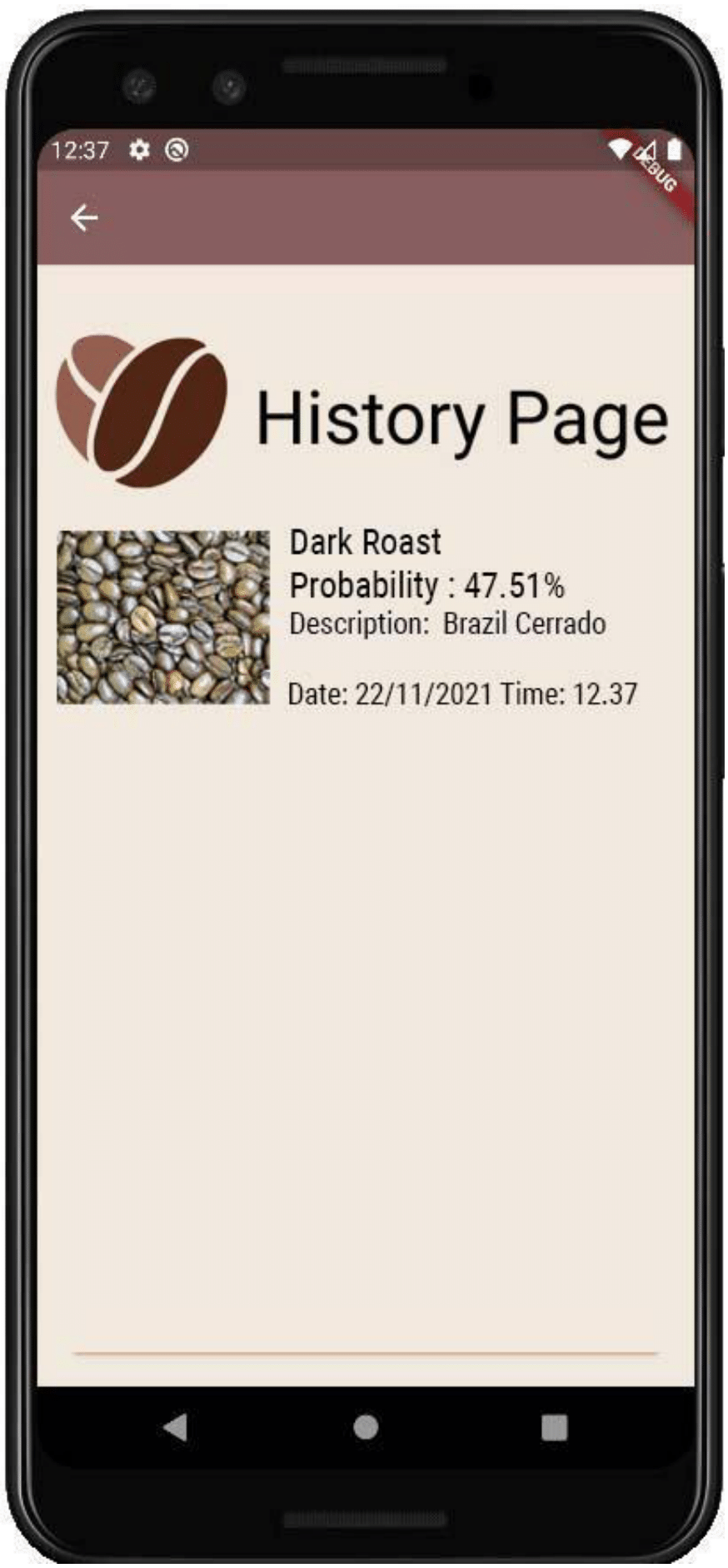}}
\caption{ History page of the “Coffee Roast Intelligence” mobile application that shows all the history of coffee beans that were saved by the user. accuracy including date and time. }
\end{figure}

\section{Conclusion}
“Coffee Roast Intelligence” is an Android application platform that is based on research related to the degree of roasted coffee bean classification using machine learning. This application helps provide better convenience and precision in the roasting of coffee beans for baristas and enterprises in the coffee shop. The main purpose of this application is to detect the color and shape of coffee beans by photographing them while they are being roasted and then uploading a photo. The color of the coffee beans changes from green (unroasted) to darker browns (dark roast). This program can notify users the level of roast, the coffee beans are at, by displaying text, as well as the percent of probability of class prediction. In addition, users can provide a description and keep a record of the results of roasting prediction.

\section{Future Development}
A disadvantage of this study is that the dataset cannot control the coffee beans’ origin. Various factors can influence the color and form of coffee beans. As a result, errors in real use may occur. A dataset of coffee beans from the same provider must be accessible in order to continue developing this project. This will aid in the prediction of outcomes’ efficiency and correctness. This project will be able to accept a wider variety of inputs by adding noise.

\section{Acknowledgement}
This study is sponsored by School of Information Technol- ogy at King Mongkut’s University of Technology Thonburi. They gave us a lot of useful advise and graciously enabled us to get all of the information we needed to finish  the  “Coffee Roast Intelligence” project. They also offered the system infrastructures for our project’s testing prior to its implementation on site.

\vspace{12pt}
\color{red}

\end{document}